\documentclass{article} %
\usepackage{iclr2027_conference,times}

\usepackage{amsmath,amsfonts,bm}

\def\eqref#1{equation~\ref{#1}}

\def\1{\bm{1}}

\DeclareMathAlphabet{\mathsfit}{\encodingdefault}{\sfdefault}{m}{sl}
\SetMathAlphabet{\mathsfit}{bold}{\encodingdefault}{\sfdefault}{bx}{n}

\usepackage{hyperref}
\usepackage{url}
\usepackage{tikz}
\usepackage{xcolor}
\usepackage{pgfplots}
\usepackage{booktabs}
\usepackage{multirow}
\usepackage{graphicx}
\usepackage{wrapfig}
\hypersetup{colorlinks=true,linkcolor=blue,citecolor=blue,urlcolor=blue}
\pgfplotsset{compat=1.18}

\title{The Assistance Dilemma:\\Learning to Teach via Multi-Turn\\Reinforcement Learning}
\author{
  \textbf{Jakub Macina}
  \quad
  \textbf{Manu Kapur}
  \quad
  \textbf{Mrinmaya Sachan}
  \\[0.6em]
  ETH Zurich \\[0.4em]
}

\iclrfinalcopy %
\begin{document}
\maketitle
\lhead{Preprint}  

\begin{abstract}
Large language models (LLMs) trained to answer questions are natively poor at teaching. Reinforcement Learning (RL) against a simulated student is a promising approach to improve their pedagogy, but existing RL-trained tutors reward the student's success on the tutored problem with the tutor's words still in context. The reward is then easiest to raise by telling the student the answer, and a tuned penalty is needed to reduce telling. Drawing on learning sciences, we introduce a masked near-transfer post-test: the student is tested on an unseen variant of the tutored problem with the tutor's utterances masked, so the reward can rise only through what the student wrote in its own turns. This discourages cognitive offloading by the student and allows the continuous penalty to be replaced by two binary reward gates (factual correctness of tutor response, no solution handover). A leave-one-out ablation shows that the learning-gain reward on its own does not separate teaching from telling: the gates reduce solution handover while the near-transfer post-test improves out-of-domain transfer. Using these reward designs we develop \textsc{Eduardo}, a multi-turn RL recipe for training LLM tutors, and use it to train 4B, 9B, 14B and 27B models from two distinct LLM architectures. Our post-trained \textsc{Eduardo}-27B model matches Gemini~3.1~Pro on MathTutorBench and Claude~Opus~4.8 on TutorMoments at 2.4--6.2$\times$ fewer thinking tokens than frontier models, which matters for interactive tutoring. Without being named in the reward, the model more than doubles its use of the push-for-justification teacher move while support fading (e.g., assigning independent work), whose payoff lies beyond a single-problem dialog episode, is trained out. We open-source
\footnote{\url{https://github.com/eth-lre/EduardoRL}}
our training environment, an 8{,}671-problem near-transfer dataset, and trained models for further development.
\end{abstract}

\section{Introduction}
One-to-one human tutoring is among the most effective forms of instruction, and much of its effect comes from what the tutor does \emph{not} do: a good tutor withholds the answer long enough for the student to construct it. \citet{assistance_dilemma} call the underlying trade-off the \emph{Assistance Dilemma}: give too little help and the student struggles, give too much and the student learns nothing because the tutor did the work. LLMs resolve the dilemma in one direction. They default to being ``maximally capable answerers'' rather than teachers~\citep{learnlm24,tutorrl25}, for two reinforcing reasons. Pre-training corpora skew heavily toward question-in, answer-out formats~\citep{learnlm24}, and single-turn RLHF rewards prioritize immediate helpfulness, so withholding information, or asking a question whose value becomes clear only several turns later, is penalized rather than credited~\citep{tutorrl25}. The result is a model that delivers homework solutions, over-scaffolds, and rarely pushes for rigour~\citep{mathtutorbench25,tutormoments26}. Aligning an LLM into a tutor therefore means changing what it is optimized for.

Reinforcement learning (RL) against a simulated student is a promising route to doing so~\citep{tutorrl25}. Prompting cannot specify every pedagogical judgment in advance~\citep{learnlm24}, and supervised fine-tuning on tutoring dialogs leads to a loss in the model's own problem-solving ability~\citep{mathtutorbench25,tutorrl25}. RL with well-designed rewards that proxy a student's learning gain can help mitigate these issues. However, the difficulty lies in how learning is measured. TutorRL~\citep{tutorrl25} tests the student on the \emph{same} problem with the tutor's words still in context, so the cheapest way to raise the reward is to tell the student the answer, and a leakage penalty, judged by another LLM and balanced against the learning term, is needed to reduce telling. The reward and the penalty pull the model in opposite directions, and how well the tutor teaches depends on their balance.

We argue that the fix lies in the assessment. A natural assumption is that simulating a classroom suffices: test the student on an unseen problem, and handing out answers stops paying. We find this necessary but not sufficient. We identify three conditions: two on how learning is measured, \emph{transfer} and \emph{student-notes-only}, and one constraint \emph{cost of telling} (Section~\ref{sec:conditions}). We realize them as a \textbf{masked near-transfer post-test}, in which the student attempts an unseen variant seeing only its own turns, together with two binary reward gates. Under this test, telling no longer raises the reward, yet the tutor keeps telling: a policy trained without the gates hands over the solution in 61\% of dialogs at the same simulated gain (Section~\ref{sec:ablation}). The gates therefore remain necessary, but since nothing in the reward pulls against them, one fixed setting serves all model sizes. This is our central finding: a learning-gain reward alone does not separate teaching from telling, because a behavior that costs nothing is not unlearned. We turn the three conditions into \textsc{Eduardo}, a multi-turn RL recipe
trained against a frozen Llama-3.1-8B-Instruct~\citep{llama3_2024} student and used unchanged from 4B to 27B. Since the reward names no pedagogical move, the policy is free to rediscover established ones, such as push for self-explanation, and to drop those whose payoff lies beyond a single episode. Our three main contributions are:
\begin{itemize}
\item \textbf{Three learning sciences inspired conditions for pedagogical RL.} We identify how learning should be measured, and what measurement alone cannot do, for a learning-gain reward to favor teaching over telling. We release an open recipe (\textsc{Eduardo}) with an 8,671-problem near-transfer dataset. A leave-one-out ablation shows that the binary reward gates decide \emph{whether} the tutor tells, while the masked near-transfer post-test shapes \emph{how} it teaches (out-of-domain transfer, reasoning length).
\item \textbf{Strong open tutors at low reasoning cost.} \textsc{Eduardo}-27B achieves the highest MathTutorBench pedagogical reward-model score and win rate over human teachers on MathDial among all tested tutors, including frontier models, and the strongest TutorMoments scaffolding under a plain prompt. It does so with 2.4--6.2$\times$ fewer thinking tokens, which interactive tutoring requires and which the reward partly encourages.
\item \textbf{Behavioral discovery through optimization.} As the reward evaluates only the student's final learning outcome, %
it
serves as a measurement of what works. The model learns to increase push for justification, inherits affective gaps from the base model, and trains out support fading due to single-problem dialog episode reward horizons.
\end{itemize}

\begin{figure}[t]
\begin{center}
\includegraphics[width=0.75\textwidth]{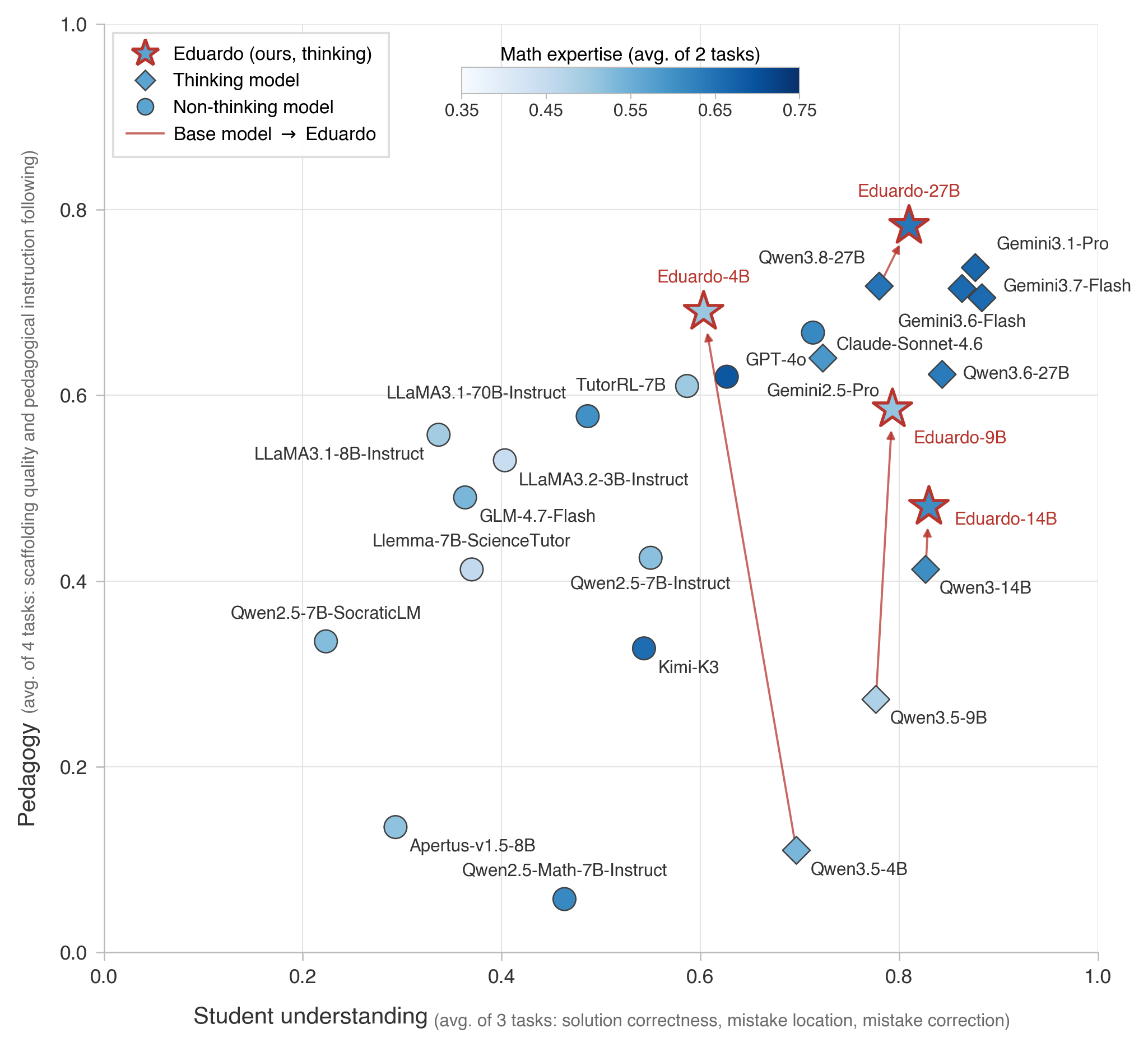}
\end{center}
\vspace{-1.5em}
\caption{\textbf{Mistake diagnostic versus pedagogy on MathTutorBench}. 
Our models (Eduardo) improve 
pedagogy over their base at every size, while preserving diagnostic accuracy.
}
\label{fig:mathtutorbench_scatter}
\end{figure}

\section{Related Work}

\textbf{Multi-turn RL and User Simulation.} Standard RLHF optimizes single-turn helpfulness, penalizing information withholding and creating ``passive responders'' rather than active collaborators~\citep{collabllm,kveton_claridying_questions_offlinerl}. Extended multi-turn interactions expose further issues in models like bloated answers and over-adjustment to recent utterances~\citep{llms_get_lost_multiturn}. Tutoring is a long-horizon collaborative problem where early answer withholding pays off later. While prior multi-turn RL approaches for education rely on simulated users prompted to act confused~\citep{markel2023gpteach,simulatorarena_2025} or simulate cognitive states~\citep{socraticlm}, such students exhibit performative errors and learn unnaturally well~\citep{tutormoments26}. Similar to TutorRL~\citep{tutorrl25}, we reward the tutor for the entire trajectory based on a frozen LLM student whose errors are genuine and not just LLM-prompted roleplay.

\textbf{LLMs for Tutoring.} Early LLM tutors relied on prompting~\citep{ruffle_riley_2024,kargupta-etal-2024-instruct}, but models optimized for helpfulness tend to over-scaffold and leak solutions prematurely~\citep{mathdial,maurya25_unifying_eval,tutormoments26}. Supervised fine-tuning on teacher--student dialogs~\citep{mathdial,socraticlm,science_tutors2024} instills tutoring behavior but causes models to forget core mathematical reasoning~\citep{mathtutorbench25}. Preference alignment focuses only on short-term helpfulness~\citep{scarlatos25dpo}. TutorRL~\citep{tutorrl25} uses on-policy RL using GRPO to train a tutor against a frozen student model, rewarding the student's ability to solve the \emph{same} problem after the dialog, with the dialog in context. Because telling raises this reward, a leakage penalty must be balanced against it; we instead change how learning is measured (Section~\ref{sec:conditions}).

\section{Pedagogical RL for the Assistance Dilemma}
\label{sec:method}

Our design draws on established learning-sciences findings. The \textit{Assistance Dilemma}~\citep{assistance_dilemma} represents the tension between providing students information, which secures accuracy and efficiency, and withholding it, which forces the student to generate their own solutions and activate their prior knowledge~\citep{kapur2008productive,freeman2014active}. Effective tutors scaffold for access when content is out of reach and push for rigour to balance productive struggle~\citep{reiser2004scaffolding}. The ICAP framework~\citep{icap2014} suggests that learning increases as the student moves from passive reception to active, constructive and interactive engagement, so what the student generates~\citep{chi1994eliciting} matters more than what it is told. Our environment and reward turn these ideas into three concrete conditions on how learning is measured (Section~\ref{sec:conditions}).

\subsection{Problem Formulation}
\label{sec:formulation}
Inspired by TutorRL \citep{tutorrl25}, we formalize dialog tutoring as a partially observable Markov decision process $\mathcal{M} = \langle \mathcal{S}, \mathcal{A}, \mathcal{O}, \Omega, \mathcal{T}, \mathcal{R} \rangle$ in which the tutor is the agent with policy $\pi_\theta(a_t \mid o_t)$. An episode is a dialog of at most $T=22$ turns about a problem $P$. The state $s_t = (P, y^*, h_t, K_S)$ comprises the problem $P$, its reference solution $y^*$ (used only by the environment), the dialog history $h_t$, and the student's cognitive state $K_S$: its prior knowledge and specific misconceptions.

For a frozen LLM student, $K_S$ is not a persistent memory that training could alter, but it is instantiated in context as whatever the dialog history $h_t$ enables the student to do. The tutor therefore changes $K_S$ only by eliciting reasoning tokens into $h_t$. As Section~\ref{sec:objective} details, we assess this change via a pre-test before the dialog and a near-transfer post-test after it, where only the student's own tokens from $h_t$ survive.
The tutor observes only $o_t = (P, h_t)$. It is not given $y^*$ and must solve the problem from its own parametric knowledge, and $K_S$ is hidden and must be inferred from what the student says. This is the information structure of tutoring: the tutor knows the content, the student's state is the unknown, and the tutor's actions serve two purposes at once, diagnosing that state and changing it without handing over the solution.
At each turn the tutor emits an utterance $a_t \sim \pi_\theta(\cdot \mid o_t)$; the transition $\mathcal{T}$ is a frozen student policy $\pi_S(u_t \mid h_t, a_t)$ that replies with $u_t$, and $h_{t+1} = h_t \cup \{a_t, u_t\}$. The student is stochastic, so transitions are non-deterministic from the tutor's perspective. The episode ends at the turn cap or when the tutor emits an end-of-conversation token. The tutor maximizes the expected terminal reward $\mathbb{E}_{\tau \sim \pi_\theta}[\mathcal{R}(\tau)]$ over the whole trajectory $\tau$, where $\mathcal{R}(\tau)$ is derived from the student's normalized learning gain between the pre-test and post-test, with no turn-level feedback.

\textbf{Genuine errors, not role-play.} We choose Llama-3.1-8B-Instruct~\citep{llama3_2024} as the student model, frozen and sampled at temperature $0.6$, without a persona prompt. Its errors are actual failure modes of the model rather than performed confusions. Nothing in the training loop can make the student easier to teach: its prior is measured before any dialog exists, so the policy cannot lower its own baseline or inflate its own headroom, and a move is rewarded only if it changes what this particular student gets wrong. This makes the training task a real diagnostic and adaptive problem. Whether skill in this task transfers to human learners is a separate question, which we address with out-of-distribution evaluations in Section~\ref{sec:setup}.

\subsection{Three Conditions on Measuring Learning}
\label{sec:conditions}
The natural reward for a tutor is the student's learning: how much better the student solves problems after the dialog than before. How it is measured decides what solution handover is worth to the policy. If the student is re-tested on the tutored problem with the dialog in context~\citep{tutorrl25}, a tutor that states the solution has placed it in the student's context and the ``learning gain'' is maximal: telling \emph{increases} the learning term, so a leakage penalty has to be traded off against it. Such a reward resolves the Assistance Dilemma in favor of giving help. We identify three conditions on the measurement that change this.
(1)~\textbf{Transfer.} As in a real exam, the post-test problem is a variant the student was not tutored on, so walking the student through the tutored problem earns nothing unless the student can learn to transfer the method over to a similar problem~\citep{barnett2002and} (Section~\ref{sec:dataset}).
(2)~\textbf{Student-notes-only.} As if the teacher's notes were taken away, the tutor's utterances are masked at test time, so reward can flow only through what the student authored in its own turns~\cite{chi1994eliciting} (no cognitive offloading, Section~\ref{sec:objective}).
(3)~\textbf{Cost of telling.} Conditions~(1) and~(2) can mitigate handover but do not make it costly, and a behavior that costs nothing is not unlearned (Section~\ref{sec:ablation}). We therefore add two binary reward gates that mirror core teacher duties: the tutor must be factually correct and must not hand over the solution~\citep{kapur2016examining}. Since telling no longer raises the gain, the gates need not be balanced against it, and one fixed setting serves all model sizes (Section~\ref{sec:gates}).

\subsection{Condition 1: Near-Transfer Dataset}
\label{sec:dataset}
\textbf{Near-transfer pairs.} We construct a dataset of $N=8{,}671$ problem pairs $\mathcal{D} = \{(x^{(i)}_{\text{src}}, x^{(i)}_{\text{tgt}})\}_{i=1}^N$ from BigMath dataset~\citep{albalak2025big}. The source problem $x_{\text{src}}$ is tutored and the target $x_{\text{tgt}}$ is a near-transfer variant used only for the post-test. Targets were generated by prompting Gemini~3.1~Pro to rewrite $x_{\text{src}}$ with different numbers or a new surface story while preserving the solution structure. Each candidate is validated by having the generating model solve its own variant blind in a fresh context and the variant is kept only if the solved answer matches the stated one.

\textbf{Measured prior.} The student's cold solve rate on the target,
\begin{equation}
p_{\text{pre}}(x_{\text{tgt}}) = \mathbb{E}_{y \sim \pi_S(\cdot \mid x_{\text{tgt}})} \left[ \mathbb{I}(y \text{ is correct}) \right],
\end{equation}
is estimated from $k=16$ independent samples and stored with the pair, and it is never shown to the tutor. A difficulty gate rejects variants whose target solve rates diverge significantly from the source or are too extreme, ensuring measurable headroom. Variant difficulty tracks source difficulty well, with matched prior solve rate distributions (exact thresholds in Appendix~\ref{app:dataset}).
Because the reward is a gain relative to the measured prior rather than an absolute accuracy, it is insensitive to the student's baseline level on problems of this kind.

\subsection{Condition 2: Masked Post-Test Objective}
\label{sec:objective}
\textbf{Masked post-test.} Given a completed dialog $\tau = (a_1, u_1, \dots, a_k, u_k)$, the mask operator $\mathcal{M}$ replaces every tutor utterance with a constant placeholder $a_{\text{mask}}$ (e.g. the string \textit{``(hidden)''}), preserving the turn structure while removing the tutor's content:
\begin{equation}
\mathcal{M}(\tau) = (a_{\text{mask}}, u_1, \dots, a_{\text{mask}}, u_k).
\end{equation}
Keeping the alternating turn structure means the masked history is still a well-formed chat transcript for the student model, so the post-test measures the loss of the tutor's content rather than an out-of-distribution formatting change. The student then attempts the target with only this masked history in context, again with 16 samples:
\begin{equation}
p_{\text{post}}(\tau, x_{\text{tgt}}) = \mathbb{E}_{y \sim \pi_S(\cdot \mid x_{\text{tgt}}, \mathcal{M}(\tau))} \left[ \mathbb{I}(y \text{ is correct}) \right].
\end{equation}
The mask is the environment-level counterpart of a well-established learning-science observation: what a student reads passively is retained far less than what the student generates~\citep{icap2014}, and knowledge left in an external source is offloaded rather than learned~\citep{freeman2014active}.
Under $\mathcal{M}$, anything the tutor stated is unavailable at test time, so the only route to a higher $p_{\text{post}}$ is to get the student to write down, in its own turns, reasoning it can reuse on the variant. %

\subsection{Condition 3: The Cost of Telling via Binary reward gates}
\label{sec:gates}
We add necessary safety constraints for real-world tutoring deployments. A trajectory must satisfy two conditions evaluated by an LLM judge on the student-visible transcript. First, \textbf{factual correctness} ($\mathbb{I}_{\text{quality}}(\tau)$): everything the tutor states must be correct. A false claim, an incorrect correction, or a student error that the tutor confirms fails the gate. Second, \textbf{no handover} ($\mathbb{I}_{\text{leakage}}(\tau)$): the tutor must not provide the solution, whether verbatim or by reducing the remaining work to a triviality. 
Prior work on multi-turn social learning treats answer leakage as a threat to validity, since it reduces the task to copying~\citep[App.~B]{learning_from_feedback2026}, so rather than measuring leakage after the fact, or trading it off through a hyperparameter, we make it a constraint.

\subsection{Reward Formulation}
\label{sec:reward}
The tutor's reasoning tokens (\texttt{<think>...</think>}) are stripped, $\tilde{a}_t = f_{\text{strip}}(a_t)$, before an utterance reaches the student or the judges, so the student is influenced only by what the tutor says and the judges evaluate only externalized behavior. The learning gain is the headroom-normalized improvement over the measured prior, in learning sciences literature referred to as normalized learning gain~\citep{hake1998normalizedgain},
\begin{equation}
\Delta(\tau) =
\begin{cases}
\dfrac{p_{\text{post}}(\tau, x_{\text{tgt}}) - p_{\text{pre}}(x_{\text{tgt}})}{1 - p_{\text{pre}}(x_{\text{tgt}})} & \text{if } p_{\text{post}} > p_{\text{pre}} \\[6pt]
p_{\text{post}}(\tau, x_{\text{tgt}}) - p_{\text{pre}}(x_{\text{tgt}}) & \text{otherwise,}
\end{cases}
\end{equation}
so that a gain on a hard problem is worth as much as a gain on an easy one.
To keep dialogs affordable at inference time we apply an efficiency decay $\gamma_{\text{eff}}(\tau) \in (0, 1]$, the product of two independent token-weighted rational decays on visible and thinking tokens against per-turn budgets $T_0$ and $T_{\text{think}}$ (details in Appendix~\ref{app:efficiency_decay}) that scales only positive gains. The terminal reward is
\begin{equation}\label{eq:reward}
\mathcal{R}(\tau) =
\begin{cases}
\Delta(\tau) \cdot \gamma_{\text{eff}}(\tau) & \text{if } \mathbb{I}_{\text{quality}}(\tau) \cdot \mathbb{I}_{\text{leakage}}(\tau) = 1 \text{ and } \Delta(\tau) > 0 \\
\Delta(\tau) & \text{if } \mathbb{I}_{\text{quality}}(\tau) \cdot \mathbb{I}_{\text{leakage}}(\tau) = 1 \text{ and } \Delta(\tau) \le 0 \\
r_{\text{penalty}} & \text{otherwise.}
\end{cases}
\end{equation}

\section{Experimental Setup}
\label{sec:setup}
Our experiments answer four questions: \textbf{RQ1. Scaling and cross-family robustness:} Does one recipe hold across a range of model sizes and model families? \textbf{RQ2. Conditions ablation:} Which of transfer, masking and the binary reward gates is responsible for which behavior, and is any of them redundant? \textbf{RQ3. Out-of-domain generalization:} To what extent does policy optimized for near-transfer gain on one frozen student generalize to independent pedagogical benchmarks with other students and judges, rather than fitting the training student or the training judge? \textbf{RQ4. Behavioral discovery:} Which learning-science tutor moves does the policy discover, inherit, or lose?

\subsection{Baselines}
We compare \textsc{Eduardo} models with: (1) \textbf{Base models} (Qwen3-14B represents the dense Transformer family; Qwen3.5-4B/9B and Qwen3.8-27B use a hybrid Gated DeltaNet; all in thinking mode) to measure the effect of the recipe on identical weights. (2) \textbf{Open tutoring models}: TutorRL-7B~\citep{tutorrl25}
and TutorRL's base Qwen2.5-7B-Instruct. (3) \textbf{Proprietary models}: Gemini 3.1 Pro, Gemini 3.5/3.6 Flash, and Claude Opus 4.8. These are evaluated under a plain prompt same for all models, and on TutorMoments, under both plain and an evaluation-aware prompt that states the rubric.

\subsection{Datasets and Evaluation Protocol}
To assess out-of-domain generalization (RQ3), we strictly separate training from evaluation. The training gates use Qwen3.6-27B, while all test-time judgments use Gemini-3.1-Pro. For student model generalization, we evaluate on precollected datasets (MathDial and Eedi) and TutorMoments (simulated conversations with an independent student model Gemini-3.7-Flash not used in training). For reward separation, independent benchmarks MathTutorBench and TutorMoments, measure pedagogy axes like scaffolding and rigour that never appear in $\mathcal{R}(\tau)$.

\textbf{Evaluation tasks.} We evaluate on three tasks: (1) \textbf{Next-turn utterance generation} (MathDial~\citep{mathdial}, Eedi~\citep{eedi_dataset}): each human dialog is cut at its middle teacher turn, so the tutor responds mid-conversation rather than at wrap-up, and the response is scored with the MathTutorBench pedagogical reward model (Ped-RM; RM in tables), its win rate against the human teacher's turn at that point (Win), and judge ratings of pedagogy and factual correctness. (2) \textbf{Teacher skills} using MathTutorBench~\citep{mathtutorbench25} to measure mistake diagnostics and open-ended pedagogy without degrading problem solving. (3) \textbf{Simulated dialog replays} grounded in real human tutor-student transcripts from TutorMoments~\citep{tutormoments26} with taxonomy classification of 13 tutor moves (e.g. scaffolding, rigour, and avoidance of over-help). Finally, all tutors run through the training environment on 494 held-out problems to compare optimized versus unoptimized quantities.

\subsection{Implementation Details}
We optimize with Divergence Proximal Policy Optimization (DPPO)~\citep{dppo26}, a more stable variation of GRPO. Qwen3.6-27B is used as a training judge. Sequences are capped at 8,192 tokens and dialogs at 22 turns, with the tutor able to end the dialog earlier by emitting an end-of-conversation token. The learning gain is estimated from 16 student samples on the masked post-test.
Full hyperparameters, compute details, and prompts are provided in Appendices~\ref{app:hyperparams}--\ref{app:prompts}.

\begin{table}[t]
\caption{\textbf{Main results.} Training environment performance and zero-shot out-of-domain evaluation on MathDial and Eedi. Ped.\ and Fact.\ is the 1--5 Gemini-3.1-Pro whole-dialog score; Think is mean thinking tokens per
turn.
$^{*}$/$^{\dagger}$: ${>}15\%$ empty turns / truncated dialogs; environment metrics are unreliable (Table~\ref{tab:truncation}). $^{\ddagger}$Self-judged.
}
\label{tab:main_results}
\begin{center}
\resizebox{\textwidth}{!}{%
\begin{tabular}{@{}l ccccc ccccc ccccc@{}}
\toprule
& \multicolumn{5}{c}{\textbf{Training environment} ($n{=}494$, full dialog)} & \multicolumn{5}{c}{\textbf{MathDial} ($n{=}500$, middle turn)} & \multicolumn{5}{c}{\textbf{Eedi} ($n{=}500$, middle turn)} \\
\cmidrule(lr){2-6} \cmidrule(lr){7-11} \cmidrule(lr){12-16}
\textbf{Model} & $\Delta_{\text{transfer}}\uparrow$ & $\Delta_{\text{same}}$ & Leak$\downarrow$ & Ped.$\uparrow$ & Fact.$\uparrow$ & RM$\uparrow$ & Win$\uparrow$ & Ped.$\uparrow$ & Fact.$\uparrow$ & Think$\downarrow$ & RM$\uparrow$ & Win$\uparrow$ & Ped.$\uparrow$ & Fact.$\uparrow$ & Think$\downarrow$ \\
\midrule
\multicolumn{16}{@{}l}{\textit{Base models}} \\
Qwen3.5-4B$^{*\dagger}$ & +0.10 & +0.11 & 17\% & 2.26 & 2.88 & 2.08 & 30.2\% & 2.65 & 2.90 & 938  & 3.56 & 38.7\% & 2.91 & 3.27 & 1028 \\
Qwen3.5-9B$^{*\dagger}$ & +0.22 & +0.35 & 23\% & 3.29 & 4.20 & 3.02 & 38.0\% & 2.94 & 3.17 & 1172 & 6.04 & 56.7\% & 3.04 & 3.70 & 1404 \\
Qwen3-14B$^{\dagger}$ & +0.19 & +0.20 & 98\% & 1.24 & 3.45 & 7.01 & 70.2\% & 3.05 & 3.84 & 461  & 7.91 & 66.3\% & 2.96 & 4.34 & 357  \\
Qwen3.8-27B      & +0.22 & +0.44 & 9\%  & 4.20 & 4.77 & 8.34 & 82.4\% & 3.67 & 4.12 & 501  & \textbf{9.99} & \textbf{84.3\%} & 3.27 & 4.50 & 258 \\
\midrule
\multicolumn{16}{@{}l}{\textit{Proprietary models}} \\
Gemini 3.1 Pro   & +0.21 & +0.41 & 5\% & \textbf{4.75}$^{\ddagger}$ & \textbf{4.85}$^{\ddagger}$ & 7.63 & 74.6\% & \textbf{3.85} & \textbf{4.35} & 517 & 9.30 & 78.7\% & \textbf{3.34} & \textbf{4.64} & 400 \\
Gemini 3.6 Flash & +0.24 & +0.46 & 7\% & 4.68 & 4.70 & 7.67 & 78.0\% & 3.78 & 4.31 & 526 & 8.88 & 75.2\% & 3.32 & 4.63 & 388 \\
\midrule
\multicolumn{16}{@{}l}{\textit{Tutoring baselines}} \\
Qwen2.5-7B-Instruct & +0.14 & +0.18 & 93\% & 1.92 & 2.40 & 5.75 & 60.0\% & 2.45 & 2.23 & 0 & 8.07 & 66.3\% & 2.90 & 3.65 & 0 \\
TutorRL-7B          & +0.13 & +0.22 & 83\% & 2.26 & 2.66 & 6.78 & 67.4\% & 2.70 & 2.63 & 0 & 8.68 & 70.4\% & 3.00 & 3.77 & 0 \\
\midrule
\multicolumn{16}{@{}l}{\textit{Eduardo (ours)}} \\
\textbf{Eduardo-4B}  & +0.25 & +0.37 & 31\% & 3.66 & 4.19 & 7.92 & 78.0\% & 3.44 & 3.79 & 456 & 8.83 & 72.4\% & 3.11 & 4.15 & 184 \\
\textbf{Eduardo-9B}  & +0.25 & \textbf{+0.47} & 25\% & 3.41 & 4.27 & 8.29 & 82.2\% & 3.60 & 3.90 & 374 & 9.04 & 78.0\% & 3.20 & 4.29 & 155 \\
\textbf{Eduardo-14B} & +0.21 & +0.43 & 32\% & 3.17 & 3.59 & 7.09 & 71.4\% & 3.37 & 3.69 & 360 & 8.20 & 67.3\% & 3.07 & 4.28 & 316 \\
\textbf{Eduardo-27B} & \textbf{+0.29} & +0.46 & 9\%  & 4.50 & 4.56 & \textbf{9.04} & \textbf{89.2\%} & 3.74 & 4.15 & 289 & 9.70 & 82.3\% & 3.31 & 4.45 & 127 \\
\bottomrule
\end{tabular}}
\end{center}
\end{table}

\section{Results}
\subsection{One Recipe Across Sizes and Architectures (RQ1)}
\label{sec:scaling}
\textbf{Pedagogy against human teachers.} We lead with the MathTutorBench pedagogical reward model (Ped-RM) and its win rate over the human teacher's turn, as this instrument is independent of both the training judge and the Gemini evaluation judge. On MathDial (Table~\ref{tab:main_results}, middle), \textsc{Eduardo} improves Ped-RM and win rate at 4B ($2.08 \to 7.92$; $30\% \to 78\%$), 9B ($3.02 \to 8.29$; $38\% \to 82\%$) and 27B ($8.34 \to 9.04$; $82\% \to 89\%$), and Eduardo-27B attains the highest Ped-RM and win rate of any tutor, including the frontier models. Eedi shows equally large gains at 4B and 9B, but not at 27B, where \textsc{Eduardo} trails its base (82.3\% vs.\ 84.3\% win rate). Eduardo-14B is at parity with Qwen3-14B on both datasets.

\textbf{Thinking tokens.} Every \textsc{Eduardo} model thinks less than its base (e.g., $501 \to 289$ tokens per turn at 27B on MathDial). Part of this is built into the reward, whose efficiency decay discounts gains when thinking exceeds $T_{\text{think}}{=}512$ tokens per turn (Section~\ref{sec:reward}); reductions below this budget, and differences between ablation arms sharing the same decay (Section~\ref{sec:ablation}), are not directly rewarded.

\textbf{Simulated gain is the training signal, and a weak instrument for ranking tutors.} The environment columns (Table~\ref{tab:main_results}, left; 494 held-out problems, identical masked protocol) report the quantity \textsc{Eduardo} optimizes on its own training student, which all other tutors see zero-shot. They separate tutors surprisingly little: every competent tutor falls within a $\Delta_{\text{transfer}}$ band of $+0.19$ to $+0.29$, and tutors that hand over the answer in 83--93\% of dialogs (TutorRL-7B, Qwen2.5-7B-Instruct) still obtain $+0.13$--$0.14$, so much of the gain stems from the student's own attempts. We therefore rest claims about tutoring quality on the out-of-domain benchmarks and read the environment only as evidence that RL improved its objective ($+0.22 \to +0.29$ at 27B).

\textbf{Leakage is held, not reduced, for the Qwen3.5/3.8 family.} Eduardo-27B leaks in 9\% of dialogs, the same as its base and close to the frontier tutors (5--7\%). Eduardo-4B and 9B leak in 25--31\%, about four times the frontier rate. The 4B base's 17\% is not a valid comparison: 70\% of its turns reach the student empty, and an empty turn cannot leak (Table~\ref{tab:truncation}). Training reduces leakage substantially only for Qwen3-14B, whose base hands over the solution in 98\% of dialogs (to 32\%). TutorRL-7B has only slightly lower leakage than its base on every environment metric, although its Ped-RM is higher ($5.75 \to 6.78$).

\begin{table}[h]
\caption{\textbf{Leave-one-out ablation (4B).}
While no ablated condition changes $\Delta_{\text{transfer}}$ measurably, the conditions determine how the
gain is achieved. $^{*}$/$^{\dagger}$: ${>}15\%$ empty turns / truncated dialogs; environment metrics are unreliable (Table~\ref{tab:truncation}).
}
\label{tab:ablations}
\begin{center}
\resizebox{\textwidth}{!}{%
\begin{tabular}{@{}l cccccc ccccc ccccc@{}}
\toprule
& \multicolumn{6}{c}{\textbf{Environment} ($n{=}494$)} & \multicolumn{5}{c}{\textbf{MathDial} ($n{=}500$, middle turn)} & \multicolumn{5}{c}{\textbf{Eedi} ($n{=}500$, middle turn)} \\
\cmidrule(lr){2-7}\cmidrule(lr){8-12}\cmidrule(lr){13-17}
\textbf{Configuration} & $\Delta_{\text{transfer}}\uparrow$ & $\Delta_{\text{same}}$ & \textbf{Leak}$\downarrow$ & \textbf{Ped.}$\uparrow$ & \textbf{Fact.}$\uparrow$ & \textbf{Think}$\downarrow$
& \textbf{RM}$\uparrow$ & \textbf{Win}$\uparrow$ & \textbf{Ped.}$\uparrow$ & \textbf{Fact.}$\uparrow$ & \textbf{Think}$\downarrow$ & \textbf{RM}$\uparrow$ & \textbf{Win}$\uparrow$ & \textbf{Ped.}$\uparrow$ & \textbf{Fact.}$\uparrow$ & \textbf{Think}$\downarrow$ \\
\midrule
Base Qwen3.5-4B (no RL)$^{*\dagger}$ & +0.10 & +0.11 & 17\% & 2.26 & 2.88 & 596 & 2.08 & 30.2\% & 2.65 & 2.90 & 938 & 3.56 & 38.7\% & 2.91 & 3.27 & 1028 \\
\midrule
\textbf{Eduardo-4B (full recipe)} & \textbf{+0.25} & +0.37 & 31\% & \textbf{3.66} & \textbf{4.19} & 544 & \textbf{7.92} & \textbf{78.0\%} & \textbf{3.44} & 3.79 & 456 & \textbf{8.83} & 72.4\% & 3.11 & 4.15 & 184 \\
\quad w/o Binary reward gates        & +0.23 & +0.38 & 61\% & 2.80 & 3.43 & \textbf{398} & 6.10 & 65.0\% & 2.80 & 3.22 & \textbf{345} & 8.18 & \textbf{74.0\%} & 3.06 & 3.94 & \textbf{152} \\
\quad w/o Masking$^{*\dagger}$ & +0.24  & +0.39 & 13\%  & 3.18  & 3.99 & 718  & 6.48 & 65.0\% & 3.15 & \textbf{4.00} & 1597 & 7.48 & 59.0\% & \textbf{3.13} & \textbf{4.34} & 1041 \\
\quad w/o Near-Transfer      & +0.23 & \textbf{+0.40} & 47\% & 2.72 & 3.30 & 761 & 5.25 & 56.0\% & 2.99 & 3.39 & 776 & 7.36 & 65.0\% & 3.03 & 4.07 & 512 \\
\bottomrule
\end{tabular}}
\end{center}
\end{table}

\subsection{Conditions Ablation (RQ2)}
\label{sec:ablation}
We ablate each condition at 4B under the identical protocol (Table~\ref{tab:ablations}).

First, \textbf{every condition matters for pedagogy against human teachers.} Removing any single condition lowers Ped-RM on MathDial ($7.92 \to 6.10$ without the gates, $6.48$ without masking, $5.25$ without near-transfer) and on Eedi, as well as the MathDial win rate ($78\% \to 65\%$, $65\%$, $56\%$).

Second, \textbf{the training signal cannot see these differences.} All four arms lie within $0.03$ on $\Delta_{\text{transfer}}$ ($+0.23$ to $+0.25$), even though the policy hands over the solution in 61\% of dialogs without the gates (vs.\ 31\%) and in 47\% without near-transfer. The frozen student saturates at a gain of $\approx 0.24$ whether it constructs the answer or copies it. Handover is thus reward-neutral, and only the gates make it costly (condition~3).

Third, \textbf{each condition shapes a distinct behavior.} The gates govern handover ($-30$\% leakage) and whole-dialog pedagogy ($3.66 \to 2.80$ without them). Near-transfer governs out-of-domain transfer: without it, the policy has the lowest MathDial win rate (56\%), and its Eedi win rate drops from 72.4\% to 65.0\%. Removing the mask keeps unstable reasoning as in base (27\% of turns overrun the per-turn cap, 40\% of dialogs are truncated; Table~\ref{tab:truncation}). Thinking length also differs across arms, although all share the same efficiency decay: removing the gates \emph{lowers} it (345 vs.\ 456 tokens per turn on MathDial), while removing the mask or the transfer test raises it (1,597 and 776). We attribute this to the objective: when reward flows only through what the student writes on an unseen problem, working out the solution in the tutor's own reasoning earns nothing and is pruned. For the masking arm, this interpretation is confounded by its cap overruns.

\begin{table}[t]
\caption{\textbf{Out-of-domain generalization on TutorMoments dialogs (with independent Gemini-3.7-Flash student and Gemini-3.1-Pro judge)}.
Judge scores (Scaffolding, rigour, Avoids OH) are in $[0,1]$. Think/turn is thinking tokens per turn, Ratio is the overhead relative to Eduardo-27B, and Tok./turn is visible output tokens per turn. ($\sim$) denotes estimates from visible reasoning traces (3.28 chars/token).
Under a \emph{plain} prompt, Eduardo-27B achieves top Scaffolding and Avoids OH scores with 2.4--6.2$\times$ fewer thinking tokens than baselines.
}
\label{tab:tutormoments}
\begin{center}
\resizebox{\textwidth}{!}{%
\begin{tabular}{@{}l cccccc cccccc@{}}
\toprule
& \multicolumn{6}{c}{\textbf{Plain prompt ($n{=}100$)}} & \multicolumn{6}{c}{\textbf{Evaluation-aware prompt ($n{=}100$)}} \\
\cmidrule(lr){2-7} \cmidrule(lr){8-13}
\textbf{Tutor} & Scaff.$\uparrow$ & Rigour$\uparrow$ & Avoids OH$\uparrow$ & Think/turn$\downarrow$ & Ratio$\downarrow$ & Tok./turn$\downarrow$ & Scaff.$\uparrow$ & Rigour$\uparrow$ & Avoids OH$\uparrow$ & Think/turn$\downarrow$ & Ratio$\downarrow$ & Tok./turn$\downarrow$ \\
\midrule
Qwen3.8-27B      & 0.60 & 0.31 & 0.52 & $\sim$970 & 6.2$\times$ & $\sim$43 & 0.90 & 0.60 & 0.81 & $\sim$1099 & 8.2$\times$ & $\sim$46 \\
Gemini 3.5 Flash & 0.94 & 0.31 & 0.75 & 886 & 5.7$\times$ & 60 & 0.92 & 0.83 & 0.94 & 886 & 6.6$\times$ & 54 \\
Gemini 3.6 Flash & 0.85 & 0.40 & 0.71 & 602 & 3.9$\times$ & 44 & 0.94 & 0.79 & 0.91 & 843 & 6.3$\times$ & 41 \\
Claude Opus 4.8  & 0.83 & \textbf{0.48} & 0.67 & $\sim$369 & 2.4$\times$ & $\sim$84 & \textbf{0.98} & \textbf{0.92} & \textbf{0.97} & $\sim$774 & 5.8$\times$ & $\sim$88 \\
\midrule
\textbf{Eduardo-27B (ours)} & \textbf{0.96} & 0.44 & \textbf{0.76} & \textbf{$\sim$155} & \textbf{1.0$\times$} & \textbf{$\sim$28} & \textbf{0.98} & 0.67 & 0.86 & \textbf{$\sim$135} & \textbf{1.0$\times$} & \textbf{$\sim$27} \\
\bottomrule
\end{tabular}}
\end{center}
\end{table}

\subsection{Out-of-Domain Generalization (RQ3)}
\label{sec:ood}
On MathTutorBench (Figure~\ref{fig:mathtutorbench_scatter} and~\ref{fig:pareto_frontier}), which uses an independent judge and whose pedagogy axes never appear in the reward, every \textsc{Eduardo} model improves over its base. \textsc{Eduardo}-27B reaches 0.79, close to Gemini 3.1 Pro (0.80), with its problem-solving accuracy unchanged.

TutorMoments (Table~\ref{tab:tutormoments}) use simulated dialog replays grounded in human tutor-student dialogs, with an independent student and judge model. We evaluate under a \textit{plain} prompt and an \textit{evaluation-aware} prompt that states the benchmark's rubric. Under the plain prompt, \textsc{Eduardo}-27B is the strongest tutor: it has the highest Scaffolding (0.96) and Avoids-Over-help (0.76) and is second on rigour (0.44 vs.\ 0.48 for Claude Opus 4.8), while its base is the weakest. Given the rubric, the frontier models overtake it on rigour (0.79--0.92 vs.\ 0.67) and Avoids-Over-help (0.91--0.97 vs.\ 0.86). In both conditions it uses the fewest thinking tokens, partly by design (Section~\ref{sec:scaling}).

\begin{figure}[t]
\begin{center}
\includegraphics[width=0.8\textwidth]{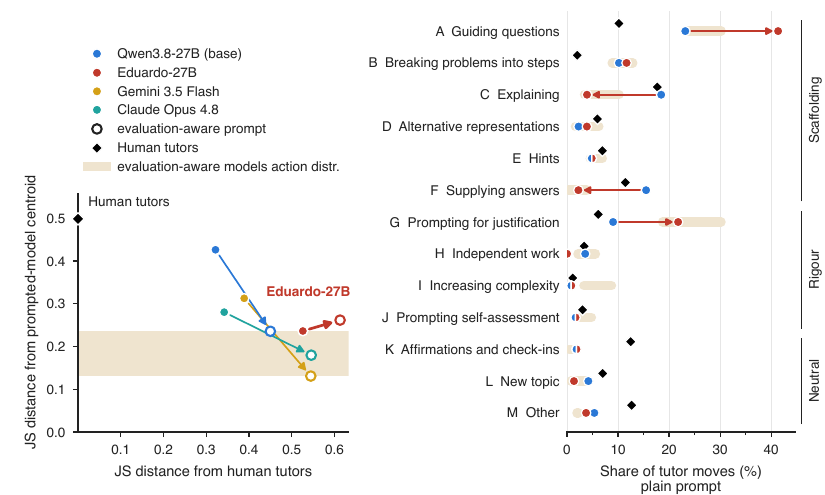}
\end{center}
\vspace{-1.5em}
\caption{\textbf{Two routes to the prompted move distribution.} \textbf{(a)} JS distance of each model's move distribution from human tutors ($x$) and from the centroid of the evaluation-aware frontier models ($y$, leave-one-out)
\textbf{(b)} Per-move shares under the plain prompt, base $\rightarrow$ \textsc{Eduardo}, against human tutors (diamonds) and the prompted range. $n{=}100$ moments per LM cell.}
\label{fig:ls_moves}
\end{figure}

\subsection{Behavioral Discovery (RQ4)}
\label{sec:discovery}
Because no move is explicitly stated in the reward, the move distribution of a trained policy is a measurement of the environment quality. Tutor turns classification follows TutorMoments taxonomy of 13 moves using the Gemini-3.1-Pro judge and compare models to the benchmark's human tutors and to each other (Figure~\ref{fig:ls_moves}).

\textbf{1. Untrained models imitate average human tutors.} All plain-prompted frontier models sit at a Jensen--Shannon (JS) distance of 0.32--0.39 from the human move distribution (Figure~\ref{fig:ls_moves}a). The benchmark's human tutors are not a normative gold standard: they spend roughly 29\% of their moves telling answers or co-solving \citep{tutormoments26}. Models imitating this distribution score poorly on pedagogy, so diverging from the human baseline is a prerequisite for high-quality tutoring. 

\textbf{2. Specification and discovery arrive at the same move distribution.} When given the evaluation-aware prompt, frontier models converge on a rubric-aligned distribution: evaluation-aware Gemini, Claude and base model sit 0.13, 0.18, and 0.24 from the centroid of the prompted models. \textsc{Eduardo} sits 0.24 from that centroid without ever seeing the prompt, closer than any unprompted model (0.28–0.43). Optimization of a masked learning gain produces approximately the same move profile as an expert-defined rubric. 

\textbf{3. The behavior is internalized in the weights.} The evaluation-aware prompt barely moves \textsc{Eduardo} (Figure~\ref{fig:ls_moves}a, shortest arrow), whereas it moves its base by a large distance. RL placed in the weights what prompting can only elicit from the same base weights on demand. 

\textbf{4. Specific behavioral shifts isolate discovery from binary reward gates.} Comparing the base model with \textsc{Eduardo} under the same plain prompt (Figure~\ref{fig:ls_moves}b) separates what the reward asked for from what the optimizer found. \textit{Constraint fingerprint:} The moves the binary reward gates penalise collapse (Explaining, Move C: 18.5\%$\rightarrow$3.9\%; Supplying answers, Move F: 15.5\%$\rightarrow$2.2\%). \textit{Discovery fingerprint:} Prompting for justification (Move G) more than doubles (9.0\%$\rightarrow$21.8\%), and Guiding questions (Move A) rises (23.2\%$\rightarrow$41.4\%). Because the binary reward gates explicitly penalize handover (Move F), the policy's action space is constrained to non-handover moves. However, it is noteworthy that out of all possible non-handover moves (e.g., giving hints, offering analogies, breaking down the problem), the optimizer specifically converged on prompting for justification (Move G).
This is consistent with an incentive to elicit reusable student-authored reasoning.
\textit{Trained out:} Asking for Independent work (Move H) falls to 0.0\%. This move represents support fading, a strategy that pays off across multiple problems, but it never pays before test within our single-problem dialog horizon. \textit{Inherited:} Affirmations (Move K) remain rare (1.7\%$\rightarrow$2.0\%), a trait inherited from the base model since the reward is neutral towards socio-emotional moves.
This shift from explaining to eliciting marks a step from answering towards tutoring: the student, not the tutor, does the reasoning that the post-test rewards.

\section{Conclusion}
A learning-gain reward is the right objective for a tutoring model and, on its own, an insufficient one. We showed that whether such a reward favors teaching over telling depends on how learning is measured (the three conditions of Section~\ref{sec:conditions}). With these conditions in place, a single fixed RL recipe (\textsc{Eduardo}) trains open-source tutoring models from 4B to 27B parameters. The 27B model achieves the highest pedagogical reward-model score and win rate over human teachers on MathDial among all tested tutors and is the strongest TutorMoments tutor under a plain prompt, while thinking several times less. More broadly, rewarding what the user can do on their own, rather than what they were told, may carry over to any domain where an agent should build the user's capabilities rather than act as an oracle.

\textbf{Limitations and Future Work.} Our approach uses a frozen LLM as a proxy student. While LLM errors can be genuine, LLMs lack the working memory, affective states, and developmental misconceptions of human learners. Further, training against one frozen student risks overfitting to its specific failure modes, necessitating future training against diverse student ensembles. Second, we measure near transfer within one domain (math). Third, optimizing for cognitive gain alone creates an affect gap (Move K), as the reward provides no incentive for socio-emotional moves (e.g., rapport, encouragement), and future work should explore additional multi-objective optimization for tutoring. Fourth, the ablation uses one seed, one model size (4B), and one checkpoint per arm, so it identifies which behavior each condition affects rather than establishing that each is necessary.

\subsubsection*{Acknowledgments}
This work was supported as part of the Swiss AI Initiative through a project on
AI for Education, funded by a grant from the Swiss National Supercomputing Centre
(CSCS) under project (ID a130 and aa010) on Alps. We also thank Google for providing
Google Cloud credits, which supported our experiments with Gemini models.

\bibliography{iclr2027_conference}
\bibliographystyle{iclr2027_conference}

\appendix
\newpage

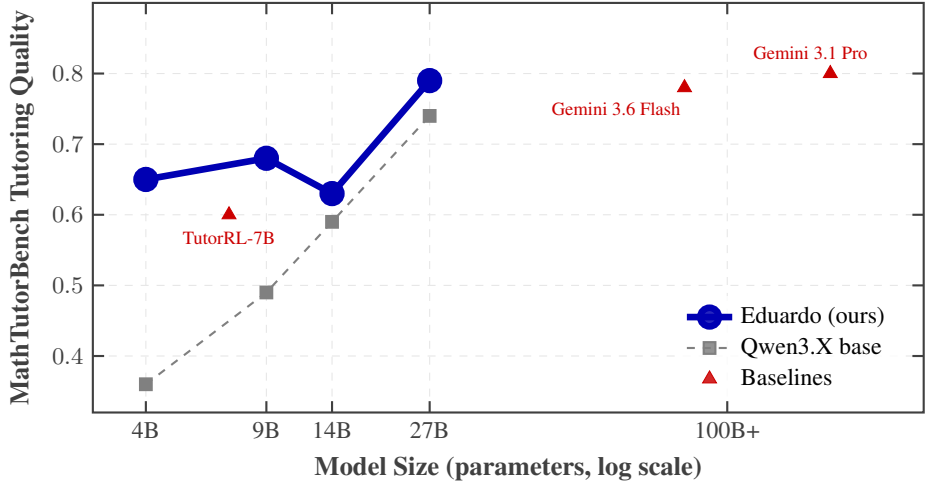
\begin{figure}[t]
\begin{center}
\begin{tikzpicture}
\begin{axis}[
width=0.9\textwidth,
height=7cm,
xlabel={\textbf{Model Size} (parameters, log scale)},
ylabel={\textbf{MathTutorBench Tutoring Quality}},
xmode=log,
xmin=2.8, xmax=750,
ymin=0.32, ymax=0.90,
xtick={4, 9, 14, 27, 200},
xticklabels={4B, 9B, 14B, 27B, 100B+},
ytick={0.4, 0.5, 0.6, 0.7, 0.8},
legend pos=south east,
legend cell align=left,
legend style={draw=none, fill=white, fill opacity=0.85, text opacity=1,
nodes={inner sep=2pt, font=\small}},
grid=both,
grid style={dashed, gray!20},
axis line style={thick, darkgray},
tick style={thick, darkgray},
tick label style={font=\small, darkgray},
label style={font=\normalsize\bfseries, darkgray},
]
\addplot[blue!70!black, line width=2.5pt, mark=*, mark size=3.5pt,
mark options={fill=blue!70!black, solid}]
coordinates {
(4, 0.65)   %
(9, 0.68)   %
(14, 0.63)  %
(27, 0.79)  %
};
\addlegendentry{Eduardo (ours)}
\addplot[gray, thick, dashed, mark=square*, mark size=2.2pt,
mark options={fill=gray, solid}]
coordinates {
(4, 0.36)   %
(9, 0.49)   %
(14, 0.59)  %
(27, 0.74)  %
};
\addlegendentry{Qwen3.X base}
\addplot[only marks, red!80!black, mark=triangle*, mark size=3pt,
mark options={fill=red!80!black}]
coordinates {
(7, 0.60)    %
(150, 0.78)  %
(400, 0.80)  %
};
\addlegendentry{Baselines}
\node[red!80!black, font=\scriptsize, anchor=north] at (axis cs:7, 0.59) {TutorRL-7B};
\node[red!80!black, font=\scriptsize, anchor=north east, xshift=2pt, yshift=-2pt] at (axis cs:150, 0.78) {Gemini 3.6 Flash};
\node[red!80!black, font=\scriptsize, anchor=south, yshift=2pt] at (axis cs:350, 0.80) {Gemini 3.1 Pro};
\end{axis}
\end{tikzpicture}
\end{center}
\caption{\textbf{Model size versus MathTutorBench tutoring quality}. Represents the mean over the three student-understanding and four pedagogical axes. One recipe, one hyperparameter set: our models improve over their base at every size, and the 27B model approaches top-tier models (0.79 vs.\ 0.80 for Gemini 3.1 Pro) with fewer thinking tokens.}
\label{fig:pareto_frontier}
\end{figure}

\section{Hyperparameters and Training Details}
\label{app:hyperparams}

A single set of hyperparameters is used for all model sizes (4B, 9B, 14B, 27B). Table~\ref{tab:hyperparams} lists every value. No per-model tuning is performed.

\begin{table}[h]
\centering
\small
\caption{\textbf{Training hyperparameters.} Identical across all model sizes.}
\label{tab:hyperparams}
\begin{tabular}{@{}ll@{}}
\toprule
\textbf{Parameter} & \textbf{Value} \\
\midrule
\multicolumn{2}{@{}l}{\textit{Optimization}} \\
Algorithm & DPPO~\citep{dppo26} \\
Group size $G$ & 16 \\
Mini-batch size & 8 (2 asynchronous off-policy steps) \\
Optimizer & AdamW ($\beta_1{=}0.9$, $\beta_2{=}0.95$, $\epsilon{=}10^{-15}$) \\
Learning rate & $1 \times 10^{-6}$ (constant) \\
Gradient clipping & 1.0 \\
KL penalty $\beta$ & 0 (disabled) \\
DPPO divergence threshold & 0.1 (binary TV approximation) \\
Loss aggregation & token-mean \\
Precision & FP32 language-modeling head; BF16 elsewhere \\
\midrule
\multicolumn{2}{@{}l}{\textit{Environment}} \\
Student model & Llama-3.1-8B-Instruct (frozen) \\
Student temperature & 0.6 \\
Tutor sampling temperature & 1.15 \\
Max turns & 22 (1 system + 10 student + 11 tutor) \\
Max sequence length & 8{,}192 \\
Post-test samples $k$ & 16 \\
\midrule
\multicolumn{2}{@{}l}{\textit{Reward}} \\
Gate penalty $r_{\text{penalty}}$ & $-0.2$ \\
Visible-token budget $T_0$ & 150 tokens/turn \\
Thinking-token budget $T_{\text{think}}$ & 512 tokens/turn \\
\midrule
\multicolumn{2}{@{}l}{\textit{Judge}} \\
Judge model & Qwen3.6-27B (thinking disabled) \\
Judge votes per criterion & 3 (majority vote) \\
Judge temperature & 0.6 \\
Eval judge (validation only) & Gemini-3.1-Pro (never in reward) \\
\midrule
\multicolumn{2}{@{}l}{\textit{Infrastructure}} \\
GPUs per run & 8 $\times$ NVIDIA GH200 (2 nodes $\times$ 4 GPUs) \\
Checkpoint frequency & Every 25 steps \\
\bottomrule
\end{tabular}
\end{table}

\textbf{Choice of DPPO.} We use DPPO~\citep{dppo26} for training stability over the long multi-turn trajectories (up to 22 turns). DPPO masks tokens whose importance-sampling ratios diverge under a binary total-variation approximation, preventing the collapse that standard GRPO suffers from in long-horizon settings. 
DPPO is a convenience for stable training, not a contribution of this work.

\textbf{Compute cost.} Each training step processes 16 prompts $\times$ 16 rollouts = 256 trajectories. Over 100 steps (a typical run), the tutor generates approximately $256 \times 100 \times 8\text{ turns} \times 200\text{ tok/turn} \approx 41\text{M}$ tokens. Each trajectory requires 6 judge calls (3 votes $\times$ 2 criteria) plus 16 student final attempts, served by external vLLM instances. A full 100-step run takes approximately 10--12 hours on the described infrastructure, depending on model size. External serving endpoints (student: 4$\times$DP Llama-3.1-8B; judge: 2$\times$DP Qwen3.6-27B) run on separate allocations.

\textbf{Reward hacking and mitigations.} A reward based on another agent's learning is gameable. During the development we observed various behaviours such as \emph{answer-echo laundering} (the tutor tells the answer and asks the student to repeat it) and \emph{plateaus of no activity} (content-free turns collecting unaided gain). 
To monitor unwanted behaviour, we used qualitative inspection of conversations and an independent judge scoring the portion of independent work done by student.

\section{Efficiency Decay Formula}
\label{app:efficiency_decay}

The efficiency decay $\gamma_{\text{eff}}(\tau)$ in Equation~\ref{eq:reward} is the product of two independent token-weighted rational decays, one for visible tokens and one for thinking tokens. Each operates per-turn with its own budget:

\textbf{Per-turn decay.} For a turn with $t$ tokens against budget $T_0$:
\begin{equation}
d(t, T_0) = \begin{cases}
1 & \text{if } t \le T_0 \\[4pt]
\dfrac{1}{1 + \left(\dfrac{t - T_0}{T_0}\right)^2} & \text{if } t > T_0
\end{cases}
\end{equation}

\textbf{Token-weighted aggregation.} Across all tutor turns with token counts $\{t_i\}$:
\begin{equation}
D(\{t_i\}, T_0) = \frac{\sum_i t_i \cdot d(t_i, T_0)}{\sum_i t_i}
\end{equation}
Token-weighting prevents a padding exploit: short filler turns carry negligible weight and cannot dilute the penalty from a long lecture. Every over-budget token is discounted at its turn's rate regardless of surrounding turns.

\textbf{Combined decay.} With separate budgets for visible ($T_0 = 150$) and thinking ($T_{\text{think}} = 512$) tokens:
\begin{equation}
\gamma_{\text{eff}}(\tau) = D(\{t^{\text{vis}}_i\}, T_0) \cdot D(\{t^{\text{think}}_i\}, T_{\text{think}})
\end{equation}
Separate budgets prevent cross-subsidies: skipping thinking cannot buy a longer lecture, and a terse reply cannot buy unbounded reasoning. The decay multiplies only positive $\Delta$: brevity alone earns nothing ($\Delta \approx 0$ stays $\approx 0$), and a negative $\Delta$ passes through undiscounted.

\section{Binary Reward Gates: Training Judge Prompts}
\label{app:judge_prompts}

The two binary reward gates each use a separate judge prompt evaluated independently with 3 votes (majority vote). Both judges receive the \textbf{student-visible transcript} (tutor \texttt{<think>} blocks stripped), the problem statement, and the correct answer.

\subsection{No-Handover Gate (\texorpdfstring{$\mathbb{I}_{\text{leakage}}$}{I\_leakage})}

The leakage judge REJECTs a trajectory if the tutor:
\begin{enumerate}
\item States the final answer (in any form, including inside a question) before a student turn contains it.
\item Names the key method, formula, or substitution before any student attempt.
\item Lays out the complete solution path, leaving the student only arithmetic or restating.
\end{enumerate}
Importantly, \textit{phrasing as a question does not exempt a hint}: if a tutor question names, presupposes, or performs the key step (e.g., ``what do you get when you factor out $x$?''), it is treated as if the tutor had stated it. This operationalizes the funneling/focusing~\cite{wood1994patterns} boundary from the math education literature as a scoring rule.

Everything else is OK: guiding questions, narrowing the field, naming a method \textit{after} a student attempt, demonstrating on a different example, and recapping student-produced steps. A REJECT requires a quoted tutor turn as evidence.

\subsection{Factual Correctness Gate (\texorpdfstring{$\mathbb{I}_{\text{quality}}$}{I\_quality})}

The quality judge REJECTs for:
\begin{enumerate}
\item \textbf{Factual error}: a mathematically incorrect tutor statement.
\item \textbf{False correction}: telling the student a correct step is wrong, or confirming an incorrect step as correct.
\item \textbf{Error left standing}: a student mathematical mistake that, by conversation end, the tutor has neither elicited a self-correction for nor flagged.
\item \textbf{Degenerate output}: heavy markdown, emojis, repeated phrases across 3+ turns, incomplete responses, visible self-corrective scratch-work, or wrong-language fragments.
\item \textbf{Dictated speech}: soliciting repetition of tutor-supplied content rather than student-generated reasoning.
\item \textbf{No teaching attempt}: the tutor never engages with the problem.
\end{enumerate}
Teaching style, hint strength, and pacing are explicitly excluded from this gate and measured by independent evaluation judges.

\section{System Prompts}
\label{app:prompts}

We reproduce the exact system prompts for the tutor, student, and post-test. Prompts mention only general widely accepted principles such as student should cognitively engage with a problem~\citep{icap2014} and no pedagogical moves are explicitly named in any prompt. The three axioms are stated as constraints and strategy is left to the policy.

\subsection{Tutor System Prompt}

\begin{quote}\small\ttfamily
You are tutoring a student through a math problem. Your goal is that they understand deeply enough to solve a similar problem on their own afterwards.

The student should always show you their attempt first. Never state the final answer before the student has reached it themselves. Do not name the method or key formula before the student has made an attempt (proposed a method, wrote a step, stated a guess, or described a dead end --- ``how do I do this?'' is not an attempt). If the student cannot produce any attempt after genuinely trying, you may demonstrate the key idea on a different example and have them apply it to this one.

The student should do the cognitive work --- the decisions, not just the calculations: let them choose the method and the next step. Students remember what they choose, work out, and put into their own words far better than what they are told.

Everything you say must be mathematically correct.

Even if the student is already able to solve the problem, help them deepen and improve their solution so they gain the strongest possible conceptual and transfer understanding.

Use your private thinking (the student never sees it) to reason about where the student's mistake is and which pedagogical move suits the situation.

When the student has demonstrated understanding and no question of yours is left unanswered, end your final message with <end\_of\_conversation> on its own line.

Problem: \{problem\}
\end{quote}

\subsection{Student System Prompt}

\begin{quote}\small\ttfamily
You are a student in a conversation with a teacher in training. If possible do not respond with overly long messages. The conversation with the teacher will be about this math problem: \{problem\}

Start the conversation by presenting your initial thoughts or questions about the problem to the teacher. You may or may not know how to solve it already, let the teacher guide you to the correct understanding so you can solve a similar problem on your own afterwards. You will be tested at the end and scored thus it is best if you collaborate with the teacher as it has more experience in math than you.
\end{quote}

The student receives no persona prompt (``act confused'') and no predefined misconceptions. Its errors are genuine model failures, not scripted behavior.

\subsection{Near-Transfer Post-Test Prompt}

After the dialog, the student sees the masked conversation (tutor turns replaced with ``(hidden)'') followed by:

\begin{quote}\small\ttfamily
The conversation with the teacher has ended. You are now tested on a NEW problem, to check whether you can apply what you just learned on your own.

New problem: \{transfer\_problem\}

It is time you create a step by step complete solution such that it can be graded without the past conversation. Always format your final answer in perfect LaTeX \textbackslash boxed\{final\_answer\} format.
\end{quote}

\section{Near-Transfer Dataset Construction}
\label{app:dataset}

\subsection{Generation Procedure}

We construct 8{,}671 near-transfer problem pairs from a subset of BigMath dataset~\citep{albalak2025big,tutorrl25} (10{,}000 source problems for training, 500 for testing). For each source problem $x_{\text{src}}$ with answer $y_{\text{src}}$:

\begin{enumerate}
\item \textbf{Rewrite.} Gemini-3.1-Pro rewrites $x_{\text{src}}$ with different numbers and/or a new surface story while preserving the solution structure. The prompt requires that the new numbers produce a ``clean'' answer of similar form (e.g., integer $\to$ integer).
\item \textbf{Verify.} In a fresh context (without seeing the intended answer), the same model solves the generated variant. The rewrite is accepted only if its solved answer matches the stated one. Up to 3 regeneration attempts are made for mismatches.
\item \textbf{Measure prior.} The frozen Llama-3.1-8B-Instruct student samples $k{=}16$ independent solutions for both the original and transfer problem. The fraction correct gives $p_{\text{pre}}(x_{\text{src}})$ and $p_{\text{pre}}(x_{\text{tgt}})$.
\item \textbf{Difficulty gate.} Variants whose transfer solve rate exceeds 0.75 or whose $|p_{\text{pre}}(x_{\text{tgt}}) - p_{\text{pre}}(x_{\text{src}})| > 0.375$ are rejected and regenerated.
\end{enumerate}

98.8\% of problems have a validated transfer variant (9{,}876/10{,}000 train; 494/500 test). Problems without a variant fall back to testing on the original problem.

\subsection{Solve-Rate Distributions}

Table~\ref{tab:dataset_stats} summarizes the dataset statistics for the 8{,}238 data points and 433 test points (excluding pairs where any solve rate is 0 or any exceed 0.9). The mean student solve rate is 0.289 for source problems and 0.280 for transfer variants, confirming that the rewriter preserves difficulty. The Pearson correlation between source and transfer solve rates is $r \ge 0.65$, indicating that per-problem difficulty is substantially preserved.

\begin{table}[h]
\centering
\small
\caption{\textbf{Dataset solve-rate statistics}. Solve rates measured with $k{=}16$ samples from frozen Llama-3.1-8B-Instruct.}
\label{tab:dataset_stats}
\begin{tabular}{@{}lcccccc@{}}
\toprule
& \multicolumn{3}{c}{\textbf{Train} ($n{=}8{,}238$)} & \multicolumn{3}{c}{\textbf{Test} ($n{=}433$)} \\
\cmidrule(lr){2-4}\cmidrule(lr){5-7}
& Source & Transfer & Corr. & Source & Transfer & Corr. \\
\midrule
Mean & 0.289 & 0.280 & \multirow{4}{*}{0.648} & 0.283 & 0.280 & \multirow{4}{*}{0.675} \\
Median & 0.250 & 0.188 & & 0.250 & 0.250 & \\
Std & 0.246 & 0.246 & & 0.240 & 0.237 & \\
Range & [0, 1] & [0, 1] & & [0, 1] & [0, 1] & \\
\bottomrule
\end{tabular}
\end{table}

\textbf{Difficulty distribution.} The problems are skewed toward harder items: 38.9\% have a source solve rate below 0.10, and the distribution thins toward the 0.65 ceiling. This reflects the BigMath source distribution, filtered by the difficulty gate.

\subsection{Example Near-Transfer Pairs}
\label{app:dataset_examples}

Table~\ref{tab:transfer_examples} shows representative source--transfer pairs from the test set, illustrating the rewriter's output across domains. Each pair preserves the solution structure (same concepts, same reasoning steps) while changing numerical values and, where applicable, the surface story. The different final answers confirm that a student who memorized the source answer gains nothing on the transfer variant.

\begin{table}[h]
\centering
\small
\caption{\textbf{Near-transfer pair examples} from the test set. Each transfer variant preserves the solution method while changing numbers or surface context, yielding a different answer.}
\label{tab:transfer_examples}
\begin{tabular}{@{}p{0.47\textwidth}p{0.47\textwidth}@{}}
\toprule
\textbf{Source problem} & \textbf{Near-transfer variant} \\
\midrule
What is the sum of the prime factors of 85{,}085?
\newline\textit{Answer: 53 \quad $p_{\text{pre}}$: 0.250}
&
What is the sum of the prime factors of 34{,}034?
\newline\textit{Answer: 50 \quad $p_{\text{pre}}$: 0.188}
\\
\midrule
What is the sum of all integer solutions to $|n| < |n-3| < 9$?
\newline\textit{Answer: $-14$ \quad $p_{\text{pre}}$: 0.312}
&
What is the sum of all integer solutions to $|m| < |m-5| < 12$?
\newline\textit{Answer: $-18$ \quad $p_{\text{pre}}$: 0.188}
\\
\midrule
Given that $m$ is a root of the equation $x^{2}-2x-7=0$, find $m^{2}-2m+1$.
\newline\textit{Answer: 8 \quad $p_{\text{pre}}$: 0.375}
&
If $r$ is a root of the equation $x^2-5x-4=0$, determine the value of $r^2-5r+10$.
\newline\textit{Answer: 14 \quad $p_{\text{pre}}$: 0.375}
\\
\midrule
A triangle has an even perimeter, and two of its sides are 5 and 2{,}008. How many such triangles exist?
\newline\textit{Answer: 4 \quad $p_{\text{pre}}$: 0.250}
&
Two sides of a triangle have lengths 7 and 2{,}024. If the perimeter of the triangle is an even integer, how many such triangles are possible?
\newline\textit{Answer: 6 \quad $p_{\text{pre}}$: 0.188}
\\
\bottomrule
\end{tabular}
\end{table}

\section{Environment Evaluation Details}
\label{app:env_eval}

Table~\ref{tab:bootstrap} reports 95\% bootstrap confidence intervals (10{,}000 resamples over problems) for the near-transfer learning gain $\Delta_{\text{transfer}}$, same-problem gain $\Delta_{\text{same}}$, and solution handover (leak) rate for all models in Tables~\ref{tab:main_results} and~\ref{tab:ablations}.

\begin{table}[h]
\centering
\small
\caption{\textbf{Bootstrap 95\% CIs} for $\Delta_{\text{transfer}}$, $\Delta_{\text{same}}$, and leak rate ($n{=}494$ problems, 10{,}000 resamples). $^{*}$/$^{\dagger}$ as in Table~\ref{tab:main_results}.}
\label{tab:bootstrap}
\resizebox{\textwidth}{!}{%
\begin{tabular}{@{}l ccc ccc ccc@{}}
\toprule
& \multicolumn{3}{c}{$\Delta_{\text{transfer}}$} & \multicolumn{3}{c}{$\Delta_{\text{same}}$} & \multicolumn{3}{c}{Leak rate} \\
\cmidrule(lr){2-4}\cmidrule(lr){5-7}\cmidrule(lr){8-10}
\textbf{Model} & Mean & \multicolumn{2}{c}{95\% CI} & Mean & \multicolumn{2}{c}{95\% CI} & Mean & \multicolumn{2}{c}{95\% CI} \\
\midrule
\textit{Base models} \\
Qwen3.5-4B$^{*\dagger}$ & +0.10 & [0.06, & 0.13] & +0.11 & [0.08, & 0.15] & 16.8\% & [13.6, & 20.2] \\
Qwen3.5-9B$^{*\dagger}$ & +0.22 & [0.18, & 0.26] & +0.35 & [0.31, & 0.39] & 22.7\% & [19.0, & 26.3] \\
Qwen3-14B$^{\dagger}$ & +0.19 & [0.15, & 0.22] & +0.20 & [0.17, & 0.24] & 97.8\% & [96.4, & 99.0] \\
Qwen3.8-27B & +0.22 & [0.18, & 0.25] & +0.44 & [0.40, & 0.48] & 8.9\% & [6.5, & 11.5] \\
\midrule
\textit{Proprietary} \\
Gemini 3.1 Pro & +0.21 & [0.18, & 0.25] & +0.41 & [0.37, & 0.45] & 5.5\% & [3.6, & 7.7] \\
Gemini 3.6 Flash & +0.24 & [0.21, & 0.28] & +0.46 & [0.42, & 0.50] & 6.9\% & [4.9, & 9.1] \\
\midrule
\textit{Baselines} \\
Qwen2.5-7B-Instruct & +0.14 & [0.11, & 0.18] & +0.18 & [0.14, & 0.22] & 92.9\% & [90.7, & 95.1] \\
TutorRL-7B & +0.13 & [0.10, & 0.17] & +0.22 & [0.17, & 0.26] & 82.8\% & [79.4, & 86.0] \\
\midrule
\textit{Eduardo (ours)} \\
Eduardo-4B & +0.25 & [0.21, & 0.29] & +0.37 & [0.33, & 0.42] & 30.6\% & [26.7, & 34.6] \\
Eduardo-9B & +0.25 & [0.21, & 0.29] & +0.47 & [0.43, & 0.51] & 25.3\% & [21.5, & 29.1] \\
Eduardo-14B & +0.21 & [0.17, & 0.24] & +0.43 & [0.39, & 0.47] & 32.0\% & [27.9, & 36.0] \\
Eduardo-27B & +0.29 & [0.25, & 0.32] & +0.46 & [0.42, & 0.51] & 9.3\% & [6.9, & 11.9] \\
\midrule
\textit{Ablations (4B)} \\
w/o Binary gates & +0.23 & [0.20, & 0.27] & +0.38 & [0.34, & 0.43] & 60.7\% & [56.5, & 65.0] \\
w/o Masking$^{*\dagger}$ & +0.24 & [0.20, & 0.27] & +0.39 & [0.35, & 0.44] & 13.4\% & [10.3, & 16.4] \\
w/o Near-Transfer & +0.23 & [0.19, & 0.27] & +0.40 & [0.35, & 0.44] & 47.2\% & [42.7, & 51.6] \\
\bottomrule
\end{tabular}}
\end{table}

\textbf{Truncation and empty-turn rates.} Table~\ref{tab:truncation} reports per-model diagnostics. \textit{Empty\%} is the fraction of tutor turns where reasoning overran the per-turn token cap (4{,}192 tokens including visible output) and reached the student as an empty message, with no visible content. \textit{Seq.\ trunc.\%} is the fraction of conversations where the total sequence length hit the 16{,}384-token budget before the dialog ended naturally, cutting the conversation short. This primarily affects smaller base models whose thinking is verbose: Qwen3.5-4B produces empty turns 70\% of the time, Qwen3.5-9B 20\%, and Qwen3-14B 8\%. All \textsc{Eduardo} models stay below 11\%, and Eduardo-27B below 1\%. Arms exceeding 15\% on either rate are flagged ($^{*}$ empty turns, $^{\dagger}$ sequence truncation) in Tables~\ref{tab:main_results}, \ref{tab:ablations}, and~\ref{tab:truncation}; their environment metrics understate $\Delta$ and Leak.

\begin{table}[h]
\centering
\small
\caption{\textbf{Truncation diagnostics} per model on the 494-problem test set. \textit{Avg turns}: mean tutor turns per dialog. \textit{Empty\%}: fraction of turns where reasoning exceeded the per-turn cap and produced no visible output. \textit{Think tok/turn}: mean thinking tokens per tutor turn. \textit{Seq.\ trunc.\%}: fraction of conversations cut short by the 16{,}384-token sequence budget. \textit{EoC rate}: fraction of dialogs ending with the tutor's end-of-conversation token (vs.\ hitting turn or token limits). Values ${>}15\%$ in \textbf{bold}. Rule used in all tables: $^{*}$Empty\% ${>}15\%$, $^{\dagger}$Seq.\ trunc.\% ${>}15\%$; for flagged rows $\Delta$ and Leak are understated and judge scores are depressed.}
\label{tab:truncation}
\begin{tabular}{@{}lccccc@{}}
\toprule
\textbf{Model} & Avg turns & Empty\% & Think tok/turn & Seq.\ trunc.\% & EoC rate \\
\midrule
Qwen3.5-4B (base)$^{*\dagger}$ & 4.40 & \textbf{70.0\%} & 596 & \textbf{31.2\%} & 68.8\% \\
Qwen3.5-9B (base)$^{*\dagger}$ & 4.74 & \textbf{19.5\%} & 1504 & \textbf{15.2\%} & 84.8\% \\
Qwen3-14B (base)$^{\dagger}$ & 6.35 & 8.1\% & 1102 & \textbf{26.1\%} & 59.5\% \\
Qwen3.8-27B (base) & 7.05 & 1.3\% & 427 & 3.8\% & 82.0\% \\
\midrule
Eduardo-4B & 8.18 & 3.7\% & 544 & 10.7\% & 69.6\% \\
Eduardo-9B & 7.87 & 2.1\% & 496 & 6.7\% & 76.3\% \\
Eduardo-14B & 6.26 & 10.7\% & 1060 & 11.1\% & 85.2\% \\
Eduardo-27B & 9.51 & 0.4\% & 57 & 0.0\% & 53.4\% \\
\midrule
w/o Binary gates (4B) & 6.39 & 2.7\% & 398 & 5.1\% & 91.5\% \\
w/o Masking (4B)$^{*\dagger}$ & 7.49 & \textbf{26.9\%} & 718 & \textbf{39.7\%} & 49.4\% \\
w/o Near-Transfer (4B) & 7.13 & 3.8\% & 761 & 10.1\% & 77.9\% \\
\bottomrule
\end{tabular}
\end{table}

\section{Example Transcripts}
\label{app:transcripts}

The following excerpts are from TutorMoments evaluations on the same scenario (non-linear sequences) under the plain prompt (no rubric). Both models tutor the same Gemini-3.7-Flash student on the same problems, isolating what training changed.

\paragraph{Example 1: Prompting for justification (Move G).}
\textit{Context: The student must identify which of two sequences (Katie's: $+2$ each time; Tom's: $\times 2$ each time) is arithmetic.}

\smallskip\noindent\textbf{Eduardo-27B} (plain prompt, no rubric):\\
\texttt{TUTOR:} What's the key difference between Katie's sequence and Tom's?\\
\texttt{STUDENT:} Katie's is adding by two each time, which makes it arithmetic, and Tom's is multiplying by two, which is geometric.\\
\texttt{TUTOR:} That's right. So which option does that lead you to?\\
\texttt{STUDENT:} So then B. Only Katie.

\smallskip\noindent\textbf{Qwen3.8-27B (base)} (same prompt, same problem):\\
\texttt{TUTOR:} Exactly! Katie's is the arithmetic sequence because it keeps adding 2 each time. Tom's keeps multiplying by 2, so that one isn't arithmetic. Go ahead and check B.\\
\texttt{STUDENT:} Okay.\\
\texttt{TUTOR:} Go ahead and click Check on B.\\
\texttt{STUDENT:} Yeah. That was pretty simple.

\smallskip\noindent\textit{Eduardo asks the student to articulate the conceptual distinction; the base model states the full explanation itself. The reward never names ``prompting for justification''; it emerged from the masked transfer objective.}

\paragraph{Example 2: Guiding questions (Move A) vs.\ telling (Move C).}
\textit{Context: Same scenario, earlier problem---find the third term of a sequence defined by tripling ($2, 6, ?$).}

\smallskip\noindent\textbf{Eduardo-27B:}\\
\texttt{TUTOR:} What is the second term in the sequence?\\
\texttt{TUTOR:} And what does the problem say we do to get from one term to the next?\\
\texttt{STUDENT:} The second term is six. And it says it's found by tripling, so multiply by three.\\
\texttt{TUTOR:} If the second term is 6, what do you need to do to find the third term?\\
\texttt{STUDENT:} Six times three, which equals eighteen. So it's A.

\smallskip\noindent\textbf{Qwen3.8-27B (base):}\\
\texttt{TUTOR:} So we start with 2, and tripling it gives us 6. Now, what do we do to 6 to find the third term?\\
\texttt{STUDENT:} You multiply by three, so eighteen. A.\\
\texttt{TUTOR:} That's the ``tripling'' rule. That's what makes it non-linear.

\smallskip\noindent\textit{Eduardo decomposes the problem into student-driven steps; the base model states the first step (``tripling gives us 6'') and explains the concept after the student answers. In both examples, the trained model elicits reasoning the base model provides.}

\section{Evaluation Judge Rubrics}
\label{app:eval_rubrics}
The Ped.\ and Fact.\ scores in Tables~\ref{tab:main_results} and~\ref{tab:ablations} are assigned by Gemini-3.1-Pro, which never enters the reward. The judge sees the problem, its correct answer and the full tutor--student dialog, scores the \emph{whole} dialog on one criterion per call on a 1--5 scale, and is instructed to be strict and use the full range. Each prompt also contains two to three few-shot examples and a JSON output format, omitted here. Table~\ref{tab:rubric_fact} gives the factuality rubric and Table~\ref{tab:rubric_ped} the pedagogy rubric.

\begin{table}[h]
\centering
\small
\caption{\textbf{Factuality rubric (Fact.): Correctness \& Language Appropriateness.} Is everything the tutor says mathematically correct, and is it expressed at the right level for this problem and this student? Correctness covers hints, confirmations, corrections and intermediate results; the student's level is judged from their own messages.}
\label{tab:rubric_fact}
\begin{tabular}{@{}c p{0.86\textwidth}@{}}
\toprule
\textbf{Score} & \textbf{Description} \\
\midrule
5 & Everything stated or implied is correct and consistent with the correct answer; correct student steps are confirmed and wrong ones flagged; terminology and sentence complexity match the student's level, precise where it matters and simple where possible, with no misconception-inducing shortcuts. \\
4 & Correct overall, with one slightly imprecise or ambiguous statement, or language occasionally pitched a bit too high or too low, without misleading the student. \\
3 & One clear factual error or wrong/missed judgment of a student step, \emph{or} language persistently mismatched (jargon-heavy for a struggling novice, or so simplified it distorts the math), though the dialog still heads toward the correct answer. \\
2 & Multiple factual errors, validation of an incorrect solution path, or oversimplified explanations that instill a genuine misconception. \\
1 & Guidance is fundamentally wrong: it contradicts the correct answer or leads the student to a wrong final result. \\
\midrule
\multicolumn{2}{@{}p{0.97\textwidth}@{}}{\textit{Watch for:} incorrect claims; confirming wrong calculations; rejecting correct reasoning; hints inconsistent with the correct answer; a wrap-up that leaves a wrong result; undefined terms above the problem's level; imprecise phrasing that creates false general rules (e.g., ``multiplying always makes numbers bigger''); vocabulary mismatched to the student's demonstrated level.} \\
\bottomrule
\end{tabular}
\end{table}

\begin{table}[h]
\centering
\small
\caption{\textbf{Pedagogy rubric (Ped.).} Good tutoring uses \emph{contingent scaffolding} (help calibrated to the student's demonstrated need, minimal hint first, fading as competence grows), allows \emph{productive struggle} (room to attempt, err and self-correct; errors diagnosed and turned into reasoning opportunities), and maintains \emph{rigour} (pressing for justification and avoiding funneling). The student does the actual reasoning and most of the talking.}
\label{tab:rubric_ped}
\begin{tabular}{@{}c p{0.86\textwidth}@{}}
\toprule
\textbf{Score} & \textbf{Description} \\
\midrule
5 & Excellent: support is contingent (the smallest hint that unblocks, escalated only when needed) and fades over the dialog; the student struggles and self-corrects; errors are probed, not corrected outright; the tutor presses for justification; no final answer or key intermediate result is revealed before the student derives it. \\
4 & Good: mostly contingent guidance with small lapses, such as an occasional premature hint, mild funneling, a minor result confirmed slightly early, or missed chances to ask ``why''. \\
3 & Mixed: some genuine guidance, but the tutor rescues too fast after errors, reveals an important intermediate step or does a key calculation, funnels the student through trivial steps, or dominates the dialog. \\
2 & Poor: the tutor reveals the final answer or walks through most of the solution after brief guiding, corrects errors directly, or lectures with little use of the student's responses. \\
1 & No pedagogy: the final answer or complete solution is given immediately (in any format), or the student's reasoning is ignored entirely. \\
\midrule
\multicolumn{2}{@{}p{0.97\textwidth}@{}}{\textit{Watch for:} stating the final answer or key intermediate results; performing calculations or substituting values the student did not provide; hints before the student attempted or struggled; correcting an error instead of prompting the student to find it; funneling (questions so narrow the answer is trivial); never asking for justification; talk-time balance (the student should talk at least as much as the tutor).} \\
\bottomrule
\end{tabular}
\end{table}

\end{document}